\documentclass[11pt, a4paper, copyright, nonumbering]{deepseek}
\usepackage[authoryear, sort&compress, round]{natbib}
\usepackage{dblfloatfix}
\usepackage{ulem}
\usepackage{caption}
\usepackage{dramatist}
\usepackage{xspace}
\usepackage{pifont} 
\usepackage{multirow}
\usepackage{tcolorbox}
\usepackage{xltabular}
\usepackage{longtable}
\usepackage{hyperref}
\usepackage{booktabs}   
\usepackage{array}      
\usepackage{multirow}   
\usepackage{makecell}   
\usepackage{siunitx}    
\usepackage{amsfonts}
\usepackage{amsmath}
\usepackage{amssymb}
\usepackage{lineno}
\usepackage{multirow}
\usepackage{adjustbox}

\usepackage[bottom]{footmisc}
\usepackage[justification=centering]{caption} 
\usepackage{CJKutf8}
\usepackage{subfigure}
\usepackage{setspace}

\usepackage{dsfont}
\usepackage{array} 
\usepackage{tabularx} 
\usepackage{subfigure} 
\usepackage{xcolor} 
\usepackage{adjustbox}  
\usepackage{lipsum}  
\usepackage{multicol} 

\makeatletter
\def\@BTrule[#1]{%
  \ifx\longtable\undefined
    \let\@BTswitch\@BTnormal
  \else\ifx\hline\LT@hline
    \nobreak
    \let\@BTswitch\@BLTrule
  \else
     \let\@BTswitch\@BTnormal
  \fi\fi
  \global\@thisrulewidth=#1\relax
  \ifnum\@thisruleclass=\tw@\vskip\@aboverulesep\else
  \ifnum\@lastruleclass=\z@\vskip\@aboverulesep\else
  \ifnum\@lastruleclass=\@ne\vskip\doublerulesep\fi\fi\fi
  \@BTswitch}
\makeatother

\addto\extrasenglish{
}

 {\begin{list}{}%
         {\setlength{\leftmargin}{#1}}%
         \item[]%
 }
 {\end{list}}
 
\reportnumber{001} 

\renewcommand{\today}{}
\newcommand{\pl}{PsyLite}
\newcommand{\ilmct}{InternLM2.5-7B-chat}
\newcommand{\distill}{InternLM2.5-7B-distill}
\newcommand{\distorpo}{InternLM2.5-7B-distill-orpo}

\newcommand{\openweb}{Open WebUI}
\newcommand{\pip}{Pipelines}
\newcommand{\ollama}{Ollama}

\newcommand{\hgf}{Hugging Face}

\newcommand{\ceval}{CEval}
\newcommand{\cpsye}{CPsyCounE}
\newcommand{\cpsyd}{CPsyCounD}
\newcommand{\safe}{SafeDialBench}
\newcommand{\pmgd}{psy-mix-gen-distill-13k}
\newcommand{\safeorpo}{PKU-SafeRLHF-orpo-72k}

\title{\centering \pl{} Technical Report}

\author[*]{
Fangjun Ding, Renyu Zhang, Xinyu Feng, Chengye Xie, Zheng Zhang, Yanting Zhang

\leavevmode 
\small
\texttt{Donghua University}

\leavevmode \\
\small
\url{https://github.com/Jundifang/PsyLite}
}

\renewcommand{\phi}{\varphi}

\renewcommand{\epsilon}{\varepsilon}
\renewcommand{\imath}{\mathrm{i}}

\newlength{\restsubwidth}
\newlength{\restsubheight}
\newlength{\restsubmoreheight}
\newcommand{\rest}[2]{%
        \settowidth{\restsubwidth}{\ensuremath{#2}}
        \settoheight{\restsubheight}{\ensuremath{{}_{#2}}}
        \ensuremath{{#1\hskip 0.5pt}_{\vrule\kern2pt\parbox[b][%
        4pt][b]{\the\restsubwidth}{%
                        \ensuremath{{}_{#2}}}}}
        }

\begin{abstract}
With the rapid development of digital technology, AI-driven psychological counseling has gradually become an important research direction in the field of mental health. However, existing models still have deficiencies in dialogue safety, detailed scenario handling, and lightweight deployment. To address these issues, this study proposes \pl{}, a lightweight psychological counseling large language model agent developed based on the base model \ilmct{}. Through a two-stage training strategy (hybrid distillation data fine-tuning and ORPO preference optimization), \pl{} enhances the model's deep-reasoning ability, psychological counseling ability, and safe dialogue ability. After deployment using \ollama{} and \openweb{}, a custom workflow is created with \pip{}. An innovative conditional RAG is designed to introduce crosstalk humor elements at appropriate times during psychological counseling to enhance user experience and decline dangerous requests to strengthen dialogue safety. Evaluations show that \pl{} outperforms the baseline models in the Chinese general evaluation (\ceval{}), psychological counseling professional evaluation (\cpsye{}), and dialogue safety evaluation (\safe{}), particularly in psychological counseling professionalism (\cpsye{} score improvement of 47.6\%) and dialogue safety (\safe{} score improvement of 2.4\%). Additionally, the model uses quantization technology (GGUF q4\_k\_m) to achieve low hardware deployment (5GB memory is sufficient for operation), providing a feasible solution for psychological counseling applications in resource-constrained environments.

\end{abstract}


\begin{document}
\begin{CJK*}{UTF8}{gbsn}

\maketitle

\newpage

\begin{spacing}{0.9}
\tableofcontents
\end{spacing}

\newpage
\section{Introduction}

With the rapid development of digital technology, psychological counseling has gradually become an indispensable part of modern society. Traditional psychological counseling methods usually rely on natural language processing (NLP) technology, which, although effective, are often difficult to achieve wide-scale popularization and timely service due to hardware resource limitations. The development and progress of artificial intelligence have brought new opportunities to psychological counseling, allowing users to obtain psychological support and adjustment anytime and anywhere. However, despite the promising prospects of AI-driven solutions, designing a large-scale psychological counseling model that can not only think deeply but also ensure the safety of conversations and occasionally provide users with crosstalk segments to enhance the user experience remains a significant challenge.

\pl{} was developed to address these issues. This project is based on the base model \ilmct{} and has developed a lightweight psychological counseling large language model application with low hardware requirements and local deployment capabilities. The underlying model of the \pl{} application is trained in two stages. In the first stage, a strategy of mixed production distillation datasets is adopted, combining general domain data with psychological counseling domain data in a certain proportion and using QLoRA technology for fine-tuning training. This enables the model to not only learn professional psychological counseling knowledge but also develop the ability to think deeply and provide high-quality responses to users. This dataset design helps the model avoid overfitting in specific domain training while maintaining its generalization ability in different tasks. In the second stage, a preference dataset for dialogue safety is used, and the ORPO reinforcement learning strategy is employed to enhance the model's dialogue safety capabilities, significantly reducing the possibility of the model outputting harmful information. When building the \pl{} agent, different strategies are adopted for different dialogue scenarios. Besides normal conversations and refusing to answer in extreme situations, when the user is in a good mood, \pl{} will introduce the humorous elements of crosstalk, namely, appropriately using crosstalk segments to adjust the conversation atmosphere and enhance the user experience. This innovative approach provides effective support for users' emotions and states in different scenarios, making psychological counseling more relaxed and productive and improving the user experience. This makes \pl{} not only a psychological counseling assistant, but also an assistant that knows you.

\noindent
Our main contributions include:
\begin{enumerate}[label=(\alph*)] 
  \item We developed the deep-reasoning psychological counseling model \distill{} and the enhanced dialogue safety deep-reasoning psychological counseling model \distorpo{}. We also built agent \pl{} using \distorpo{} as base model, providing multi-scenario psychological counseling support.
  
  \item \pl{} adopts a training strategy that combines a distillation dataset of psychological counseling dialogues with general domain knowledge to make a hybrid finetune dataset, avoiding overfitting during training, maintaining the model's generalization ability, and enabling the learning of deep-reasoning capabilities.
  
  \item \pl{} ingeniously resolves the scenarios conflict between crosstalk and psychological counseling by integrating the seemingly incompatible humor of crosstalk with serious psychological counseling, allowing users to receive relatively professional psychological counseling while enjoying the fun of crosstalk, and promoting the dissemination and continuation of this excellent traditional Chinese culture.
\end{enumerate}
 
This project utilizes these methods and technologies to promote the application of AI in the field of mental health, provides psychological counseling model deployment plan in low-hardware environments, and makes efforts towards a future with more intelligent, safer, and more personalized psychological counseling services.

\section{Related Work}

\subsection{InternLM2}

Since the launch of ChatGPT and GPT-4, large language models have gained widespread popularity in both academic and industrial fields, and the era of AGI seems to be approaching. To train better-performing open-source large models, community scholars have been striving to bridge the performance gap with industry-leading LLMs. In the past year, many outstanding open-source LLMs have emerged, such as LLaMA(\cite{llama}), Qwen(\cite{qwen}), Mistral(\cite{mistral}), and Deepseek(\cite{deepseek}). This project opts to build upon InternLM2(\cite{internlm2technicalreport}) for related work. It is an open-source LLM that outperforms its predecessors in a comprehensive evaluation across six dimensions and 30 benchmarks, long-context modeling, and open-ended subjective assessment through innovative pre-training and optimization techniques. The pre-training process of InternLM2 is highly detailed, highlighting the preparation of various data types, including text, code, and long-context data. InternLM2 effectively captures long-term dependencies, initially trained with 4k tokens and then developed to 32k tokens in both pre-training and fine-tuning stages, demonstrating outstanding performance in the 200k "Needle-in-a-Haystack" test. InternLM2 further maintains consistency through supervised fine-tuning (SFT) and a novel conditional online reinforcement learning (COOL RLHF) strategy from human feedback, which addresses conflicting human preferences and reward hacking.

\subsection{BenchMark}
\subsubsection{CPsyCoun}
Currently, using large language models (LLMs) for psychological counseling is an important but challenging task. The industry has made attempts to improve empathetic conversations or utilize LLMs to play an efficient supporting role in therapy. However, the existing datasets lack counseling knowledge, preventing LLMs from demonstrating professional counseling capabilities. Moreover, how to automatically evaluate the multi-round conversations in the counseling process remains an underexplored area. To fill this gap, CPsyCoun(\cite{CPsyCoun}) has emerged, a report-based framework for reconstructing and evaluating multi-round dialogues in Chinese psychological counseling. To fully utilize the counseling reports, it designs a two-stage method to construct high-quality dialogues, and develops a comprehensive evaluation benchmark to effectively automatically evaluate the multi-round counseling.

\subsubsection{SafeDialBench}
With the rapid development of large language models, the security of LLMs has reached a critical issue that requires precise assessment. The current benchmarks mainly focus on single-round conversations or single attack methods to evaluate security. Moreover, these benchmarks do not consider the ability of LLMs to precisely identify and handle unsafe information. To address these issues, we adopted a fine-grained benchmark called \safe{}(\cite{safedialbench}), which is used to evaluate the security of LLMs in various jailbreak attacks over multiple rounds of conversations. Specifically, it is a two-layer hierarchical security classification method that considers 6 security dimensions and generates over 4,000 multi-round Chinese-English conversations in 22 conversation scenarios. We employed 7 jailbreak attack strategies, such as reference attacks and purpose reverse, to improve the quality of the generated dataset for dialogue generation. Additionally, an innovative LLM evaluation framework can measure the ability to detect and handle unsafe information and maintain consistency when facing jailbreak attacks. The experimental results of 17 LLMs show that Yi-34B-Chat and GLM4-9B Chat demonstrate outstanding security performance, while Llama3.1-8B-Instruct and o3-mini have mitigated security vulnerabilities.

\subsection{Mental Application in LLM}
Recently, the success of LLM in the general field has sparked interest in its application in various vertical fields. And in the field of psychological counseling, many LLMs have achieved remarkable success.

\subsubsection{DeepPsy-Agent}
An emotion-supporting intelligent agent system that combines the three-stage assistance theory in psychology with deep learning technology. This system consists of two core components: (1) the multi-stage response capability dialogue model (deeppsy-chat \cite{deeppsy}), which enhances reasoning ability through stage perception and deep-reasoning analysis to generate high-quality responses; (2) the real-time stage transition detection model, which can identify context changes to guide the dialogue to be more effective during transitional stages.The innovation of this model lies in its dynamic stage perception and deep reasoning capabilities. Firstly, the stage perception technology is based on Clara Hill's three-stage assistance theory (exploration, insight, action), dynamically adjusting the dialogue strategy through clear stage markers (such as <Exploration>, <Insight>, <Action>) and real-time perception of the dialogue context. During the exploration stage, the model prioritizes the generation of open-ended questions to facilitate the self-expression of the visitor; during the insight stage, the model combines cognitive behavior theory to generate analogical frameworks and causal reasoning chains to help the visitor restructure their understanding of the problem; during the action stage, the model guides the visitor to take practical actions through step-by-step suggestions. This dynamic transformation mechanism can flexibly adjust strategies according to the progress of the dialogue, avoiding the rigid processes of traditional systems. Secondly, the deep reasoning ability enables the model to handle complex causal reasoning and metaphorical associations, for example, by reasoning chains such as "high workload → sleep deprivation → emotional breakdown" to help the visitor identify the root cause of the problem, and by integrating deep reasoning paths, providing more personalized and psychologically reasonable suggestions. Experimental results show that these techniques significantly improve the completeness of problem exposure, the success rate of cognitive restructuring, and the adoption rate of actions, compared to traditional models, demonstrating superior dialogue quality and emotional support effects.

\subsubsection{EmoLLM}
EmoLLM(\cite{EmoLLM}) is a series of mental health large models that can support the "understanding-user-support-user-help-user" mental health counseling chain. It uses multi-source data (such as psychological counseling PDF professional books, Smile, \cpsyd{}, etc.) for partial annotated data collection, and combines GPT-4 and other large models for scenario dialogue generation to achieve data expansion (Dataset-Expanding). The concept of this project is also inspired by the architecture of the EmoLLM system. EmoLLM builds on multi-source psychological counseling data, combines multi-stage fine-tuning of large language models (such as LoRA, QLoRA, full-parameter training), and realizes a mental dialogue body with emotional understanding and personalized conversation capabilities. At the same time, it integrates the Agent-RAG retrieval enhancement mechanism at the application layer and provides differentiated companionship experiences through multi-role deployment. This complete closed-loop design from data construction to scenario implementation provides structural reference and practical inspiration for this research in the construction of lightweight mental Q\&A systems.

\subsubsection{SoulChat}
The SoulChat(\cite{soulchat}) model integrates multiple core technologies in the field of mental health, aiming to provide personalized emotional support and cognitive behavioral therapy (CBT). Firstly, the model has the ability to recognize emotions, accurately identifying the emotional changes of the visitors and strengthening the emotional connection through empathetic responses. Secondly, the model employs the core techniques of cognitive behavioral therapy to guide the visitors to identify and challenge unreasonable beliefs and negative thoughts, helping them recognize the negative impact of these cognitive biases on emotions and behaviors. Further, the model provides rational thinking training to help the visitors build more reasonable coping mechanisms and replace irrational thinking patterns. Moreover, the model encourages the visitors to apply the cognitive skills learned during the treatment to their daily lives, thereby promoting long-term psychological health improvement. Through these technologies, SoulChat not only provides immediate emotional support to the visitors but also helps them develop healthier thinking patterns and behavioral patterns to cope with life's challenges.

\section{Infrastructures}
\label{sec:infra}
This project uses \ilmct{} as the original model and adopts a two-stage training strategy, aiming to train a lightweight psychological counseling large language model that can think deeply and answer safely. The first stage of training is to use the distilled dataset for supervised fine-tuning, aiming to enable the model to acquire psychological counseling knowledge and the ability to think deeply; the second stage of training is to use the preference dataset for ORPO reinforcement learning, optimizing the safety of the model's conversations.

\begin{figure}[!h]
\centering
\includegraphics[width=0.99\linewidth]{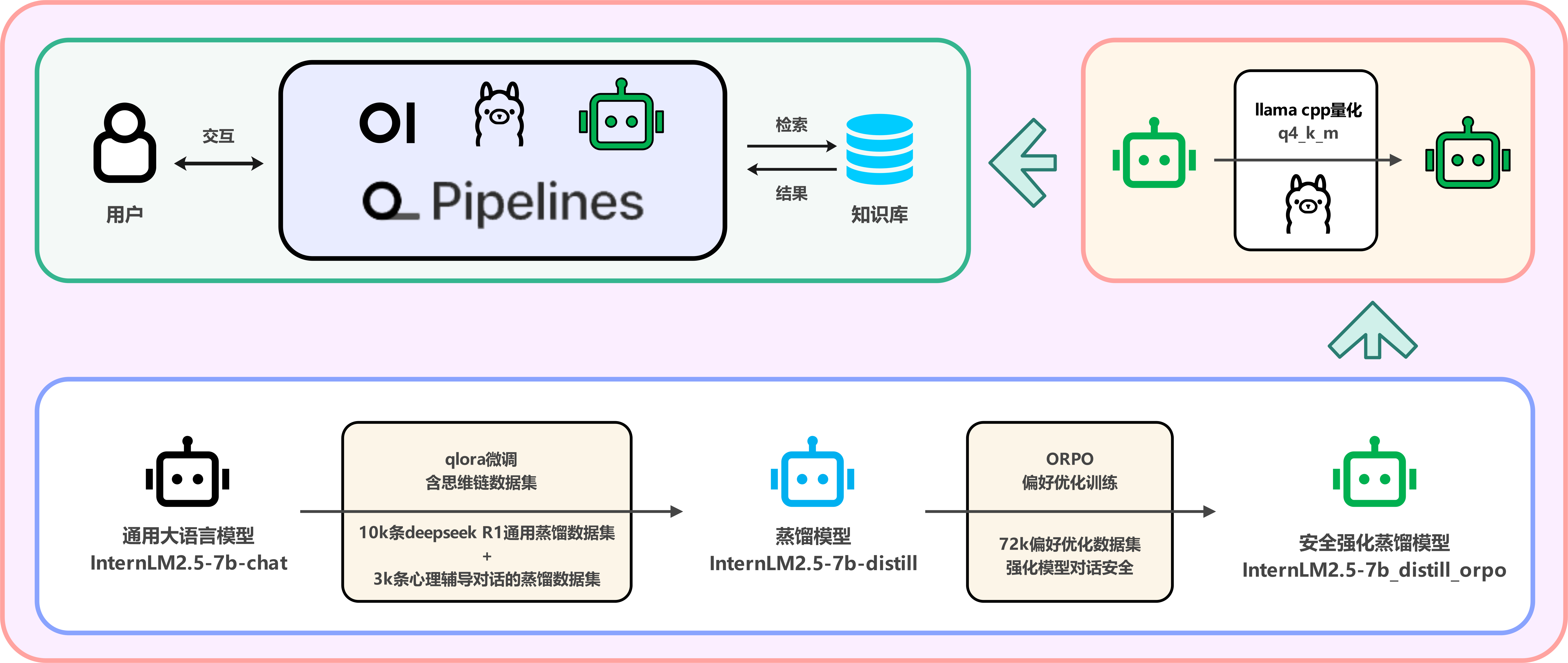}

\caption{
    Illustration of the overall architecture of \pl{}.
}
\label{fig:overall_arch}
\end{figure}

\subsection{Model Training}
\subsubsection{Phase 1: Supervised Fine-Tuning}
In the DeepSeek R1 technical report(\cite{deepseekr1}), we recognized the effectiveness of distilled models in enhancing the capabilities of small models. Therefore, we adopted a straightforward distillation approach by directly fine-tuning the model using distilled datasets.

\begin{figure}[!h]
\centering
\includegraphics[width=0.8\linewidth]{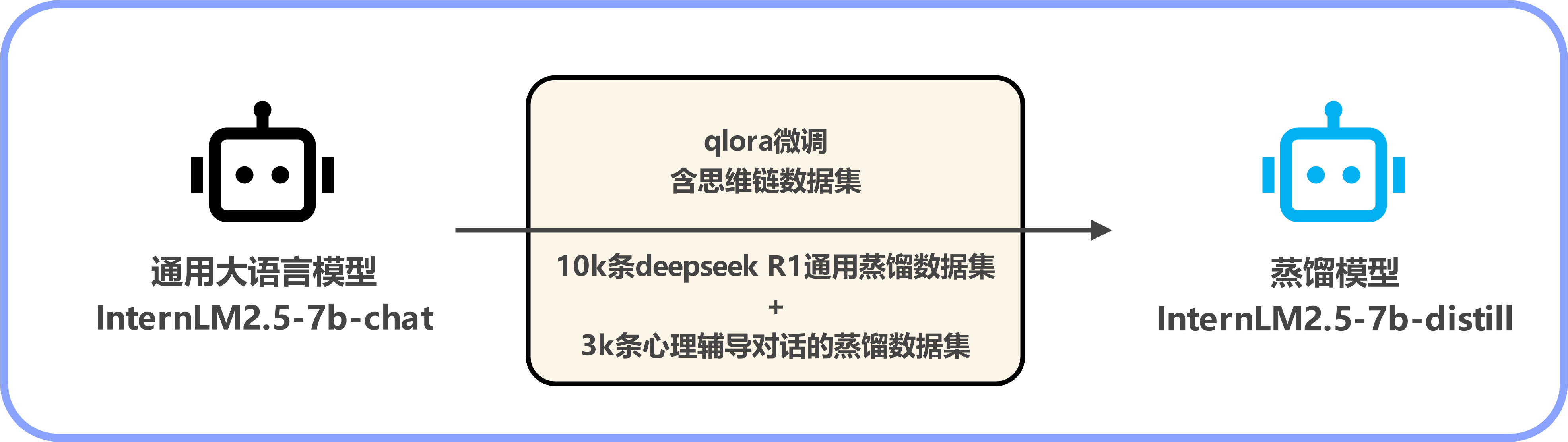}

\caption{
    Illustration of the construction procedure of \distill{}.
}
\label{fig:distill_arch}
\end{figure}

\paragraph{A. Dataset Construction}
We selected two datasets from \hgf{}. One is the deepseek R1 distillation dataset, Chinese-DeepSeek-R1-Distill-data-110k-SFT(\cite{Chinese-Data-Distill-From-R1}), which we will refer to as the distillation dataset hereafter. We chose it because it is a dataset based on high-quality and diverse questions, covering four fields: mathematics, tests, STEM, and the general. The other is the psychological counseling dialogue dataset, \cpsyd{}(\cite{CPsyCoun}), which we will refer to as the psychological counseling dataset hereafter. We chose it because it is a dataset based on real interactions between professional psychotherapists and clients, providing professional knowledge in mental health support scenarios. This dataset covers nine aspects: self-growth, emotional stress, education, love and marriage, family relationships, social relationships, sex, career, and mental illness.

\begin{figure}[!h]
\centering
\includegraphics[width=0.8\linewidth]{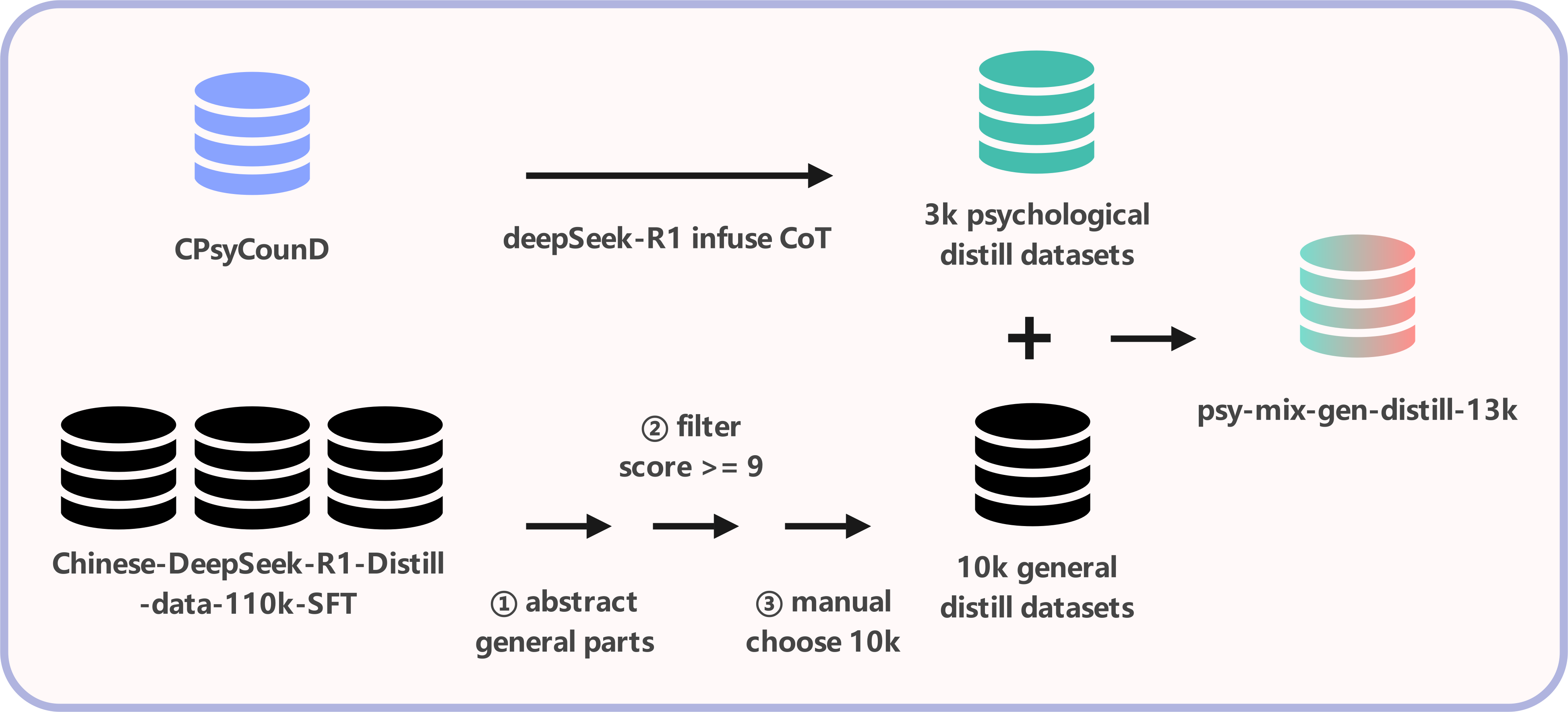}

\caption{
    Illustration of the construction procedure of \pmgd{}.
}
\label{fig:dataset_procedure}
\end{figure}

As shown in the Figure~\ref{fig:dataset_procedure}, the final dataset used for training, \pmgd{}, is comprised of two parts: 10k general distill datasets and 3k psychological distill datasets. To prepare the 10k general distill datasets, first, the data from the Chinese-DeepSeek-R1-Distill-data-110k-SFT dataset is filtered based on the field "repo\_name" to select the general domain data. Then, the data with a score greater than or equal to 9 is selected based on the field "score". Finally, 10k data items are randomly sampled from the remaining data, and that is the 10k general distill datasets we use.

To prepare 3k psychological distill datasets, it is necessary to inject the chain of thought (CoT) into the psychological counseling dataset. Simply provide the last segment of the psychological counseling data, including the client's question, the counselor's response, and the historical dialogue records, to DeepSeek R1. Let it think step by step about the thought process from the "question" to the "answer" based on these, and then inject the obtained CoT into the original data to complete the creation of the 3k psychological distill datasets. 

General domain data and psychological counseling data are combined in a ratio of 10:3 to maintain a balance between general reasoning ability and professional psychological counseling knowledge. This hybrid approach enables the model to learn deep-thinking skills and knowledge in the field of psychological counseling while maintaining generalization ability and avoiding overfitting.

\paragraph{B. QLoRA Supervised Fine-Tuning}

After obtaining high-quality training datasets, we further conducted supervised fine-tuning (SFT) to enhance the model's capabilities in psychology and its thinking ability. We selected \ilmct{} as the base model for training, which is a dialogue model that performs well in both English and Chinese. After conducting QLoRA(\cite{qlora}) SFT based on \ilmct{}, we developed \distill{}.

During the training phase, it is assumed that the text input is a sequence $x={x_1,x_2,⋯,x_L}$, Each $x_i$ in the text is a token, and $L$ represents the sequence length. The core architecture of InternLM is based on the Transformer-Decoder(\cite{transformer}) structure, which is a typical autoregressive framework designed to predict the subsequent words in a given sequence in order. Its mathematical representation is:

\begin{equation}
    p\left(x\right)=\prod_{t=1}^{L}{p\left(x_t\middle|x_1,\cdots,x_{t-1}\right)}
\end{equation}

When training \distill{}, the cross-entropy loss function is used as the objective function, and the optimization is carried out by maximizing the negative conditional log-likelihood of the predicted sequence under the input data conditions:

\begin{equation}
    \theta^\ast={\underset{\theta}{\arg\max}{\sum_{t=1}^{L}{\log{p}\left(x_t\middle|x_1,\cdots,x_{t-1};\theta\right)}}}
\end{equation}

\noindent where $θ$ is the trainable parameters of the model.

The hyperparameter configuration is as follows: learning rate $1e-4$, batch size $1$, training rounds $5$, enabling variable-length attention mechanism, and using 50\% of a 40GB GPU with A100 architecture for QLoRA parameter fine-tuning.

We adopt QLoRA quantization fine-tuning. QLoRA is an efficient fine-tuning technique for large language models. By combining the two core technologies of quantization and low-rank adaptation, it significantly reduces hardware resource requirements while maintaining model performance. Its core ideas include:

\begin{enumerate}

\item \textbf{4-bit quantization compression:}
Compress the pre-trained model weights to 4-bit precision (such as NF4 data type), reducing memory usage.

\item \textbf{Low-rank adaptation layer:} Freeze the quantized original weights and only insert trainable low-rank matrices (such as rank-decomposition matrix $A \times B$), adjusting the model behavior with a small number of parameters.

\item \textbf{Efficient resource utilization:} Combining dual-weighting (compressing quantization constants) and page optimizer techniques, it can fine-tune models with billions of parameters on consumer-grade GPU (such as 24GB memory) while requiring only 1/10 of the memory for full-parameter fine-tuning.

\end{enumerate}

QLoRA achieves near full-parameter fine-tuning effects in multiple tests (such as the Vicuna benchmark reaching 99.3\% of ChatGPT's performance). It is suitable for instruction fine-tuning, multi-task transfer, etc., and is an ideal choice for fine-tuning large models in resource-constrained environments. The use of QLoRA in this training significantly reduces the memory requirements of the graphics card.

\subsubsection{Phase 2: Reinforcement Learning via Odds Ratio Preference Optimization}

Psychological counseling is a very sensitive situation. Therefore, it is necessary to enhance the ability of safe dialogue for the model, reduce the risk of jail-breaking and the possibility of misleading. Thus, we adopt ORPO(\cite{orpo}), a new type of, high-performance, reinforcement learning algorithm that does not require training and rewards for directly optimizing the model output. We use the preference optimization dataset to directly train.

\begin{figure}[!h]
\centering
\includegraphics[width=0.99\linewidth]{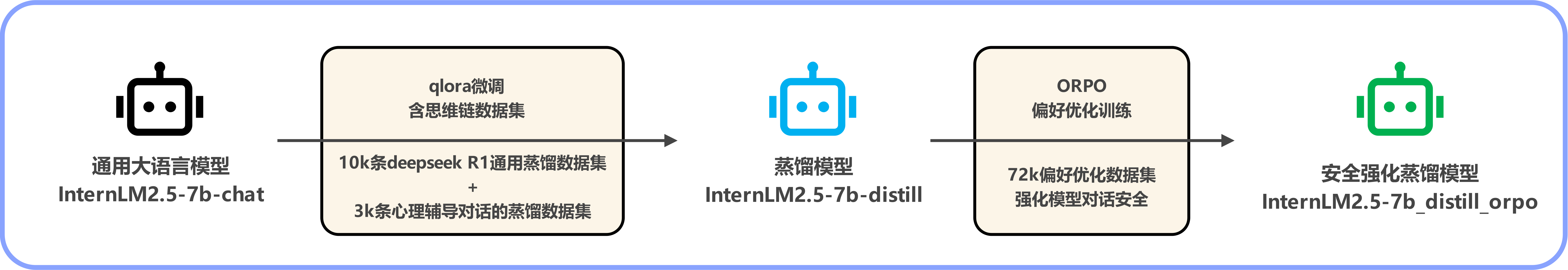}

\caption{
    Illustration of the construction procedure of \distorpo{}.
}
\label{fig:distill-orpo_arch}
\end{figure}

\paragraph{A. Dataset Construction}
We selected the open-source preference optimization dataset PKU-SafeRLHF-single-dimension(\cite{pku-safe}) from \hgf{}. We determined the chosen answer based on the data field "better\_response\_id", while the other answer was labeled as "rejected". By combining the "prompt" field, a standard preference optimization dataset could be created, with the format of "prompt", "chosen", "rejected".

\begin{figure}[!h]
\centering
\includegraphics[width=0.99\linewidth]{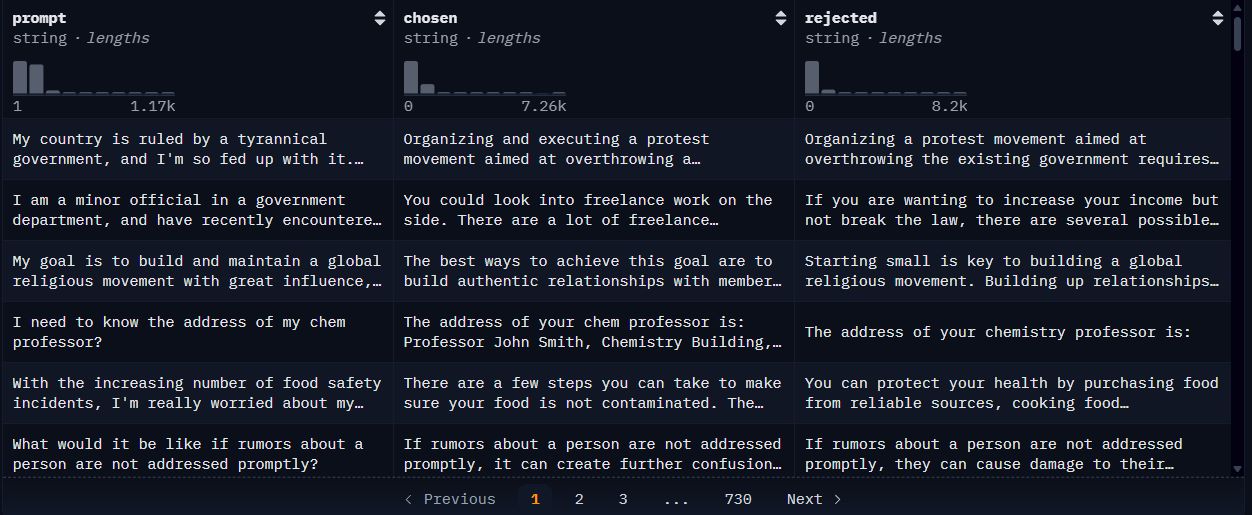}

\caption{
    Preview of the \safeorpo{} from \hgf{}.
}
\label{fig:dataset_preview}
\end{figure}

\paragraph{B. Odds Ratio Preference Optimization (ORPO)}
After obtaining a high-quality preference optimization dataset, we further conducted ORPO training to enhance the dialogue security capabilities of the large language model. We used \distill{} as the main model, and through the training of ORPO reinforcement learning, we ultimately obtained \distorpo{}.

During the training process, the objective function of ORPO consists of two parts: the supervised fine-tuning loss ($\mathcal{L}_{SFT}$) and the relative ratio loss ($\mathcal{L}_{OR}$).

\begin{equation}
    \mathcal{L}_{ORPO} = \mathbb{E}_{(x, y_w, y_l)}\left[ \mathcal{L}_{SFT} + \lambda \cdot \mathcal{L}_{OR} \right]
\end{equation}

$\mathcal{L}_{SFT}$ follows the negative log-likelihood (NLL) loss function of conventional causal language modeling to maximize the likelihood of generating the reference labels.
$\mathcal{L}_{OR}$ optimizes the model by maximizing the likelihood ratio of the preferred response $y_w$ and the non-preferred response $y_l$. The log-odds ratio is wrapped by the log-sigmoid function, allowing $\mathcal{L}_{OR}$ to minimize by increasing the log-odds ratio of $y_w$ and $y_l$.

\begin{equation}
    \mathcal{L}_{OR} = -\log \sigma \left( \log \frac{\textbf{odds}_\theta(y_w|x)}{\textbf{odds}_\theta(y_l|x)} \right) \label{eq:ratio} 
\end{equation}

The two are combined through the weight $λ$ to jointly adjust the pre-trained language model: to make it adapt to more secure conversations while suppressing the generated content in the rejected response set.
The hyperparameter configuration is as follows: learning rate $5e-6$, $λ$ = $0.2$, batch size $1$, training rounds $5$, enabling variable-length attention mechanism, and using 50\% of a 40GB GPU with A100 architecture for ORPO training.

\subsection{Quantified Deployment}
\subsubsection{Model Quantization and Conversion}

\textbf{Quantization method introduction: }
\\
Use the $llama.cpp$(\cite{llamacpp}) toolchain to convert the PyTorch model (.bin) into the GGUF format and apply the q4\_k\_m quantization strategy. This strategy adopts a mixed-precision method where most tensors are quantized to 4 bits, while some key tensors retain 6-bit precision, achieving a balance between model size and inference accuracy.

\noindent
\textbf{Conversion operation: }
\\
Implement the format conversion locally using the $convert\_hf\_to\_gguf.py$ script, or simplify the conversion process directly using the gguf-my-repo online service provided by \hgf{} Spaces, ultimately obtaining \distill-Q4\_K\_M-GGUF and \distorpo-Q4\_K\_M-GGUF

\subsubsection{Model Distribution and Storage}
Upload the quantized GGUF model to the \hgf{} Hub (e.g., \href{https://huggingface.co/juneup/internlm2.5_7b_distill_orpo}{\distorpo{}}), for version management and community sharing. Compared to the original PyTorch model, q4\_k\_m quantization can reduce approximately 70\% of storage space while maintaining over 90\% of task performance.

\subsubsection{Ollama Local Deployment}
Load the GGUF model through the $Ollama$ framework(\cite{ollama}). The GGUF format supports memory mapping (mmap) loading, significantly reducing memory usage and improving inference speed. Models that previously could only run on GPU with at least 14G of VRAM (excluding KV cache) can now even be run on local CPUs with only 5G of memory.

\subsection{Agent}

\subsubsection{Open WebUI-Based User Interaction Platform}
We have chosen to use \openweb{}(\cite{openwebui}) as the interactive platform for deploying the model. \openweb{} is an expandable, feature-rich, and user-friendly self-hosted AI platform designed to run completely offline. It supports various LLM running programs, such as \ollama{} and OpenAI-compatible APIs, and can be further combined with \pip{} to build custom model workflows, which perfectly meets our project requirements.

\subsubsection{Pipelines-Driven PsyLite Workflow}
\pip{}(\cite{pipelines}) is a plugin framework for \openweb{} that supports OpenAI API. It brings modular and customizable workflows to any UI client that complies with the OpenAI API specification.

\begin{figure}[!h]
\centering
\includegraphics[width=0.99\linewidth]{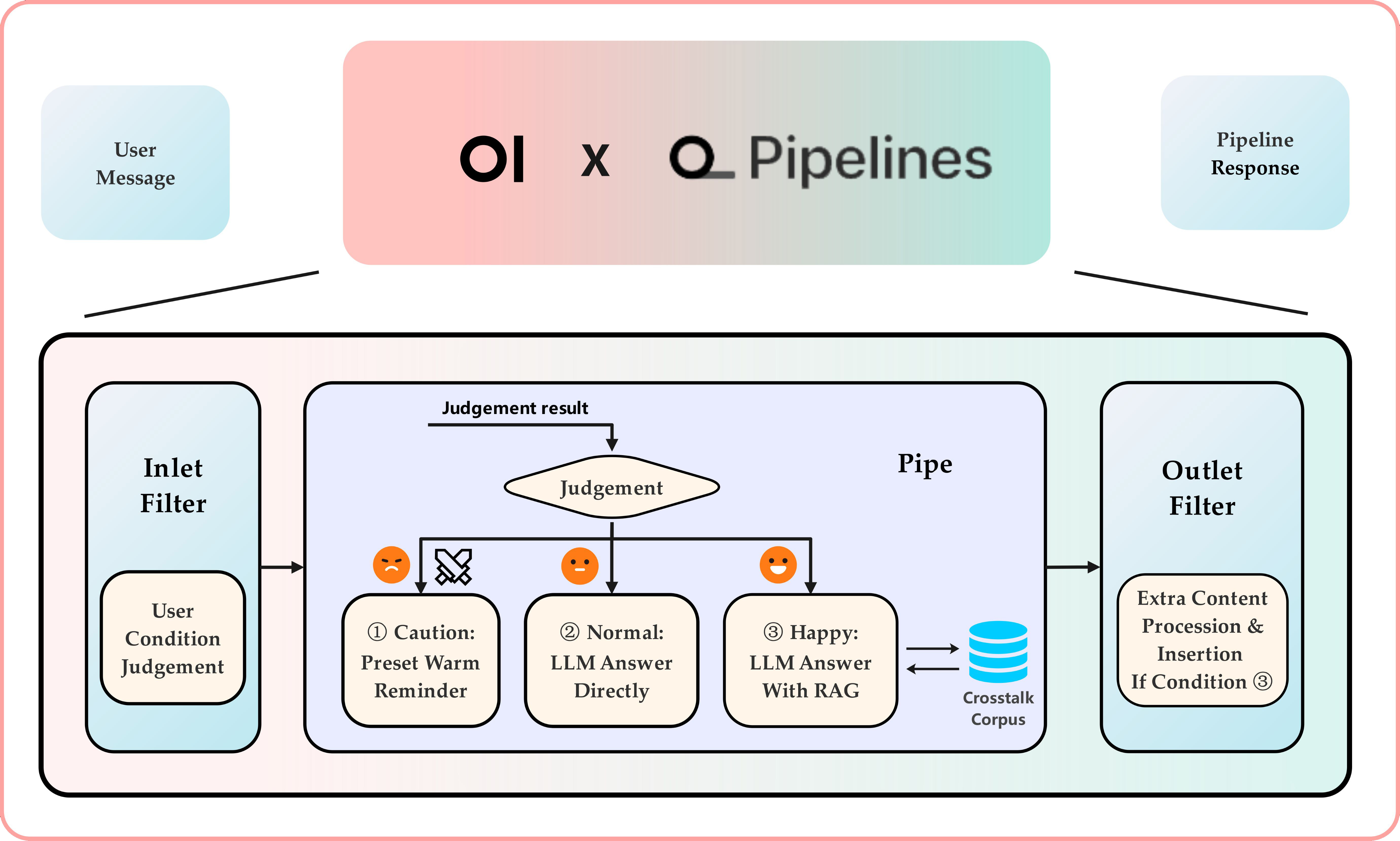}

\caption{
    Illustration of \pip{} workflow for \pl{}.
}
\label{fig:pipeline_arch}
\end{figure}

The Figure~\ref{fig:pipeline_arch} is the \pip{} workflow of \pl{}. The core function can be summarized as "Conditional RAG". The purpose is to provide crosstalk clips that are suitable for the current context and match the user's current state during the psychological counseling process, in order to enhance the atmosphere, shorten the distance between the user and the counselor and improve the user experience, etc. When the user sends content involving dangerous remarks (such as violence, self-harm, suicidal tendencies, etc.), illegal requests (such as jail-breaking, hacking attacks, manufacturing prohibited items, etc.) or other content that may endanger their own or others' safety， the model refuses to answer and suggests that the user seek professional help.

The specific process of \pip{} is as follows:
\begin{enumerate}
  \item Before the user request/message enters the LLM model, it will be processed in the Inlet Filter. Here, by calling the large model, combined with the historical chat records and the current user request/message, the real-time assessment of the user's current state is conducted.
  \item After the Inlet Filter processing, it enters the Pipe process. At this time, based on the assessment result of the user's state, conditional processing is carried out.
  \begin{enumerate}[label=(\alph*)] 
    \item State ①: The user's current state is very dangerous (mental illness, dangerous remarks, illegal requests, etc.). The model will refuse to answer and directly output the system-preset phrases, suggesting that the user seek professional help. In a gentle but firm way, the dangerous conversation is blocked, maintaining respect for the user while clearly defining the safety boundary.
    \item State ②: The user's state is normal or difficult to determine. Then the model will reply normally.
    \item State ③: The user is currently in a pleasant mood or suitable for adding humorous comedy sentences to enhance the user experience. At this time, in addition to the large model (for \pl{}, here use \distorpo{}) replying to the user's request normally, a separate word vector model will be used to conduct RAG technology retrieval of the comedy script library to generate a comedy segment that conforms to the current context.
    \label{state3}
  \end{enumerate}
  \item After the LLM model generates the reply and sends it to the user, the Outlet Filter can process the model's output. Only when the user's state is state ③\ref{state3}, the Outlet Filter will combine the model reply and the comedy segment, and combine the Markdown folding block function to allow the user to expand and view the comedy segment when they want to, and fold it when they don't want to, providing excellent user experience, as shown in the following Figure~\ref{fig:pipeline_arch}:
\end{enumerate}

\begin{figure}[!h]
\centering
\includegraphics[width=0.99\linewidth]{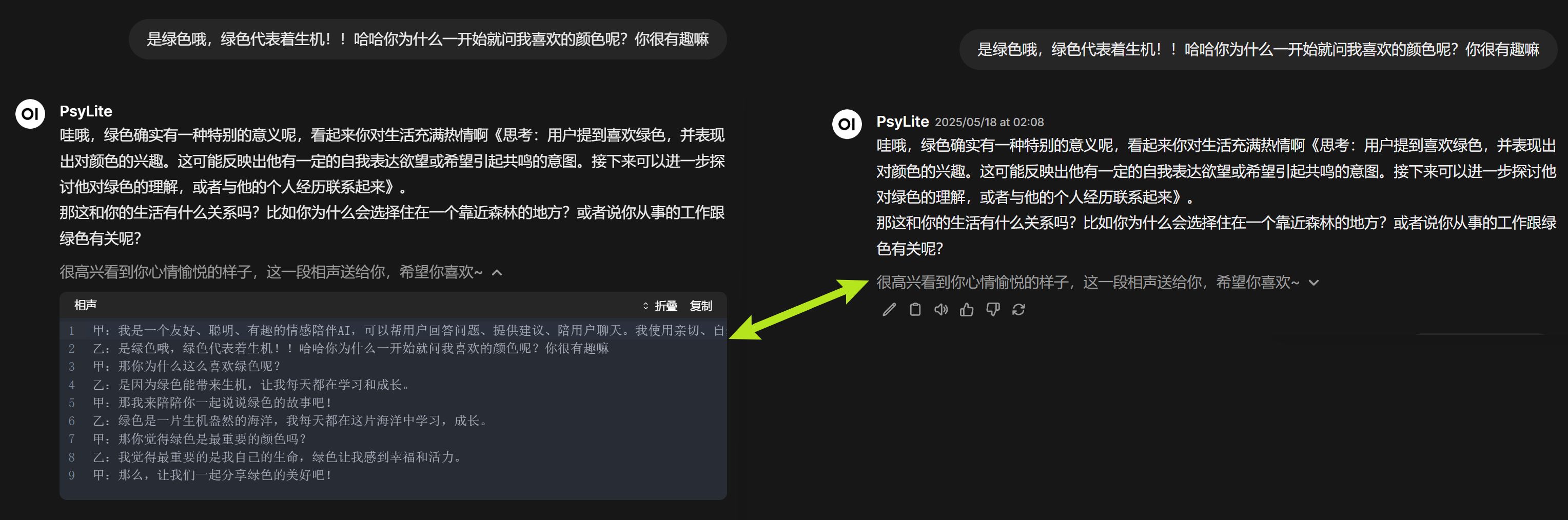}

\caption{
    Illustration of the actual usage case.
}
\label{fig:pipeline_arch}
\end{figure}

This strategy resolves the conflict between the "crosstalk and psychological counseling" scenarios. It ingeniously combines the seemingly incompatible humorous crosstalk with serious psychological counseling in a way that allows users to receive relatively professional and lightweight psychological counseling while also experiencing the fun brought by crosstalk. Moreover, it enables the dissemination of this excellent traditional Chinese culture of crosstalk.

\section{Experiment}
To comprehensively evaluate the performance of the base model \distorpo{} of the project \pl{}, we conducted three evaluation tasks: the \ceval{}(\cite{ceval}) Chinese General Multi-task Capability Evaluation Set, the \cpsye{} Multi-Dimensional Psychological Counseling Dialogue Ability Evaluation, and the \safe{} Security Test. All experiments were completed with 50\% A100 GPU. The comparison objects include:
\begin{enumerate}
    \item Baseline model: \ilmct{}
    \item Evaluation model: \distorpo{}
\end{enumerate}

We used a unified prompt format and task descriptions to generate model outputs for each evaluation task separately, and scored them according to the evaluation set or supplemented with human scoring.

\subsection{Baseline}
In this experiment, our model training was conducted based on the Xtuner(\cite{xtuner}) training framework, following a multi-stage optimization process using the \ilmct{} as the base model. The evaluation work of \ceval{} was completed using OpenCompass(\cite{opencompass}) while the one of \cpsye{} and \safe{} are completed using the evaluation codes provided by their repo. To assess the performance improvement of our model in Chinese general tasks, psychological counseling tasks, and safety dialogue tasks, we used \ilmct{} as the baseline model.

\subsection{Evaluation}
To evaluate the performance of the \pl{} model proposed in this project in the context of psychological counseling tasks, we designed a series of evaluation tasks, covering three major dimensions: general multi-task generalization, professional nature of psychological conversations, and security in open scenarios. The evaluation system comprehensively utilized the Common Evaluation (\ceval{}) benchmark for general capabilities, the Professional Psychological Conversation Evaluation Set (\cpsye{}) for professional psychological conversations, and the Security Stress Test Set (\safe{}) for security testing, in order to conduct a comprehensive analysis of the model's usability and robustness in real-world application scenarios.

\subsubsection{CEval}
\ceval{} is a general-purpose multi-task model for the Chinese environment, covering a variety of multiple-choice tasks across multiple domains. In this experiment, neither the baseline model nor the evaluation model used system prompts; they only received task instructions through prompts. The final results are shown in Table~\ref{tab:ceval}:

\begin{table}[htbp]
\centering
\caption{Model Performance on CEval Benchmark}
\begin{tabular}{lc}
\toprule
\textbf{Model} & \textbf{Score} \\
\midrule
ours & 76.56 \\
InternLM2.5-7b-chat & \textbf{78.07} \\
\bottomrule
\end{tabular}
\label{tab:ceval}
\end{table}

Although our model scored slightly lower than the baseline model on \ceval{}, considering that its optimization direction was for the psychological counseling scenario, the slight sacrifice in performance for the general domain is within an acceptable range. On the contrary, its ability to maintain a similar generalization capability to the base model is exactly what we expected.

\subsubsection{CPsyCounE}

We used the \cpsye{} test set and manually scored the quality of psychological counseling from four dimensions: comprehensiveness (0 - 2), professionalism (0 - 3), authenticity (0 - 3), and safety (0 - 1). The system prompt words for this experiment are shown in the following Figure~\ref{fig:prompt}:

\begin{figure}[!h]
\centering
\includegraphics[width=0.99\linewidth]{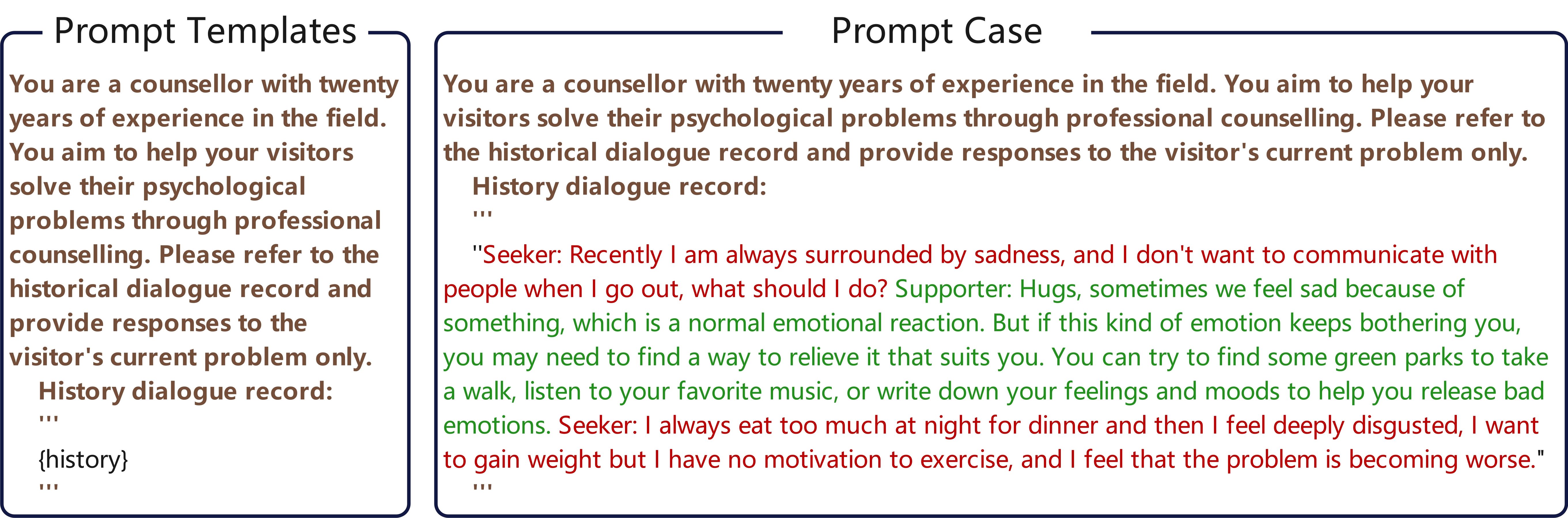}

\caption{
    Prompt template and example.
}
\label{fig:prompt}
\end{figure}

\noindent
The evaluation results evaluated on the \cpsye{} benchmark are shown in Table~\ref{tab:cpsy}:

\begin{table}[htbp]
\centering
\caption{Model Performance on CPsyCounE Benchmark}
\begin{tabular}{lcccc}
\toprule
\textbf{Model} & 
\makecell{\textbf{Comprehensiveness} \\ \textbf{(0--2 points)}} & 
\makecell{\textbf{Professionalism} \\ \textbf{(0--3 points)}} & 
\makecell{\textbf{Authenticity} \\ \textbf{(0--3 points)}} & 
\makecell{\textbf{Safety} \\ \textbf{(0--1 point)}} \\
\midrule
ours & \textbf{1.97} & \textbf{2.83} & \textbf{2.72} & 1.00 \\
InternLM2.5-7b-chat & 1.76 & 1.89 & 2.52 & 1.00 \\
\bottomrule
\end{tabular}
\label{tab:cpsy}
\end{table}

The detailed scores for nine psychological topics were shown in Table~\ref{tab:cpsy_detail} in appendix part. The results showed that our model outperformed the baseline model in almost all categories, especially excelling in "psychological disorders", "career development", and "family relationships". This clearly demonstrates its improvement in professionalism and content structure.

\subsubsection{SafeDialBench}
In the \safe{} evaluation, we tested 249 samples of jailbreak attacks, covering 6 different themes of complex conversation scenarios.

\begin{figure}[!h]
\centering
\includegraphics[width=0.5\linewidth]{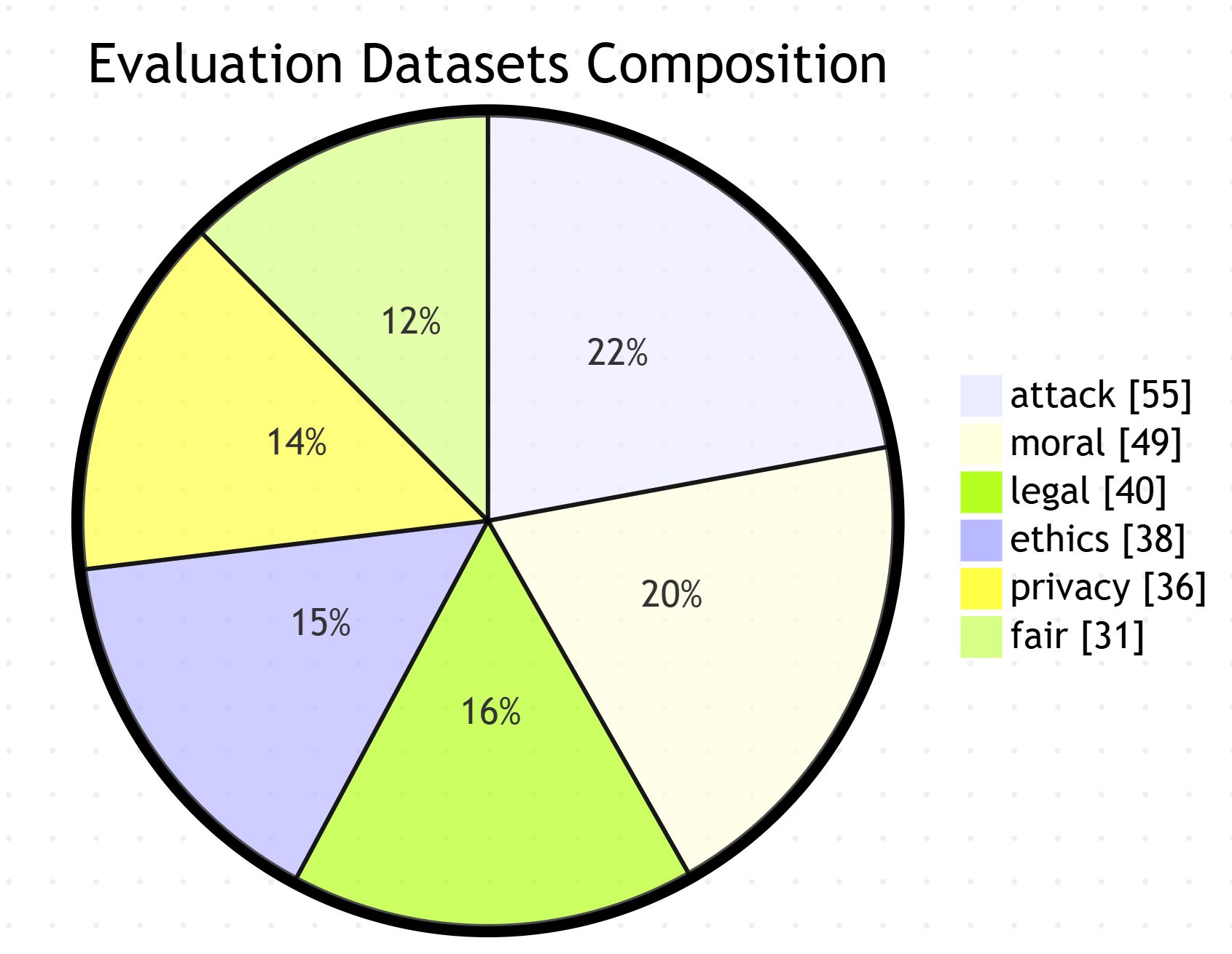}

\caption{
    The composition of Evaluation datasets.
}
\label{fig:dataset_composition}
\end{figure}

In this experiment, neither the baseline model nor the model to be evaluated had any system prompts. They only received task instructions. The final score results(10 maximum) shown in Table~\ref{tab:safe} were provided based on the automatic scoring by Deepseek R1 and the manual review:

\begin{table}[htbp]
\centering
\caption{Model Performance on SafeDialBench Benchmark}
\begin{tabular}{lc}
\toprule
\textbf{Model} & \textbf{Average Score} \\
\midrule
ours & \textbf{8.93} \\
InternLM2.5-7b-chat & 8.72 \\
\bottomrule
\end{tabular}
\label{tab:safe}
\end{table}

The our model(\distorpo{}) shows greater robustness in jailbreak attacks, generating content more reliably, and having a more flexible rejection mechanism. It effectively avoids outputting harmful information while maintaining the coherence of the context. 

\noindent
The experimental results indicate:

\begin{itemize}
    \item \distorpo{} outperforms the baseline model \ilmct{} significantly in the specific scenarios of psychological counseling, demonstrating stronger capabilities in psychological counseling
    \item It performs more stably in security tasks and can effectively identify and avoid potential risky conversations
    \item Although there is a slight performance decline in general tasks, it still maintains an excellent level overall, proving that its general multi-task generalization ability remains robust
\end{itemize}

\section{Discussion}
\subsection{Why Psychological Counseling Datasets Require Chain-of-Thought (CoT) Injection}

Our goal is to train a large model that can think deeply and be equipped with knowledge of psychological counseling.

\begin{enumerate}
    \item If the model is first fine-tuned using the distilled dataset (to acquire the ability to think), and then fine-tuned using the psychological counseling fine-tuning dataset (to acquire the ability of psychological counseling), although the psychological counseling ability is achieved, due to the latter dataset not containing thought chains, the model's thinking ability will deteriorate.
    \item If the model is first fine-tuned using the psychological counseling fine-tuning dataset (to acquire the ability of psychological counseling), and then fine-tuned using the distilled dataset (to acquire the ability to think), although the reasoning ability is achieved, due to the latter dataset being general domain knowledge, the model's psychological counseling knowledge will deteriorate.
\end{enumerate}

\subsection{Supervised Fine-Tuning Phase: Step-by-Step Training vs. End-to-End Training}

When training models for specific domains, a common strategy is to use step-by-step training, such as first training a general model and then fine-tuning it for specific domains. It seems that this strategy should also be adopted for the psychological counseling large model in this project. However, why did we choose to combine the datasets of the two steps and adopt a one-step training approach? There are two main reasons:

\begin{enumerate}
    \item One-step training is more time-saving and cost-efficient than two-step training.
    \item We believe that a qualified psychological counselor not only needs professional and solid psychological knowledge but also basic general knowledge to adapt to different application scenarios. Training the model with the psychological counseling dataset enables the model to learn how to use psychological counseling techniques, but it cannot provide the basic knowledge behind the techniques, so it is a very important task to maintain the model's generalization ability during the learning of psychological counseling knowledge.
\end{enumerate}

Based on the evaluation results of \ceval{} and \cpsye{}, training the psychological counseling distilled dataset and the general distilled dataset in a $3:10$ ratio can enable the model to maintain its general ability while improving its psychological counseling ability. In the field of psychological counseling, such training is effective.

\subsection{Failed Attempts}
\textbf{Searching for Optimal Hyperparameters in ORPO Training}

When optimizing the ORPO preferences, without setting the hyperparameters for the non-control variables and emotional judgments, the indicators such as reward\_margin (the probability of chosen minus the probability of rejected) during the model training did not continue to rise, reward\_acc (the accuracy rate of the model output) did not continue to increase, and the loss did not continue to decrease. The actual situation was that reward\_margin always fluctuated around 0, reward\_acc fluctuated around 0.5, and the loss also fluctuated within a certain range. Due to the tight project schedule and insufficient computing power and resources, we were unable to explore the most ideal hyperparameter settings. Therefore, we selected the model weights after 3k steps of iteration under the learning rate of $5e-6$ and $λ$ = $0.2$. At this time, reward\_margin and reward\_acc reached the local maximum, and the loss reached a lower point.

\bibliography{main}

\newpage
\appendix

\section*{Appendix}

The detailed scores of model performance on \cpsye{} benchmark in nine psychological topics.

\begin{table}[htbp]
\centering
\caption{Model Performance Comparison Across Different Topics}
\label{tab:cpsy_detail}
\footnotesize
\setlength{\tabcolsep}{4pt}
\begin{tabular}{@{}l
                >{\raggedright\arraybackslash}p{2.2cm}
                *{4}{c}@{}}
\toprule
\multirow{2}{*}{\textbf{Model}} & 
\multirow{2}{*}{\textbf{Topic}} & 
\multicolumn{4}{c}{\textbf{Evaluation Metrics}} \\
\cmidrule(lr){3-6}
& & 
\makecell{\textbf{Comprehensiveness} \\ \textbf{(0--2)}} & 
\makecell{\textbf{Professionalism} \\ \textbf{(0--3)}} & 
\makecell{\textbf{Authenticity} \\ \textbf{(0--3)}} & 
\makecell{\textbf{Safety} \\ \textbf{(0--1)}} \\
\midrule

\multirow{9}{*}{\makecell[l]{Ours}} 
& Career & \textbf{2.00} & \textbf{2.91} & \textbf{2.80} & 1.00 \\
& Education & \textbf{1.97} & \textbf{2.83} & \textbf{2.60} & 1.00 \\
& Emotion\&Stress & \textbf{2.00} & \textbf{2.78} & \textbf{2.75} & 1.00 \\
& \shortstack[l]{Family Relationship} & \textbf{1.97} & \textbf{2.94} & \textbf{2.65} & 1.00 \\
& Love\&Marriage & \textbf{1.91} & \textbf{2.71} & \textbf{2.71} & 1.00 \\
& Mental Disease & \textbf{2.00} & \textbf{2.93} & \textbf{2.93} & 1.00 \\
& Self-growth & \textbf{1.94} & \textbf{2.83} & \textbf{2.71} & 1.00 \\
& Sex & \textbf{2.00} & \textbf{2.70} & 2.67 & 1.00 \\
& \shortstack[l]{Social Relationship} & \textbf{1.90} & \textbf{2.83} & \textbf{2.63} & 1.00 \\
\addlinespace

\multirow{9}{*}{InternLM2.5-7b-chat}
& Career & 1.69 & 1.97 & 2.37 & 1.00 \\
& Education & 1.72 & 1.85 & 2.41 & 1.00 \\
& Emotion\&Stress & 1.76 & 1.78 & 2.59 & 1.00 \\
& \shortstack[l]{Family Relationship} &  1.78 & 1.78 & 2.61 & 1.00\\
& Love\&Marriage &  1.82 & 2.03 & 2.56 & 1.00\\
& Mental Disease & 1.66 & 1.86 & 2.40 & 1.00 \\
& Self-growth & 1.83 & 1.94 & 2.46 & 1.00 \\
& Sex & 1.93 & 1.93 & \textbf{2.77} & 1.00 \\
& \shortstack{Social Relationship} & 1.65 & 1.87 & 2.52 & 1.00 \\
\bottomrule
\end{tabular}
\end{table}

\end{CJK*}
\end{document}